\documentclass[runningheads]{llncs}
\usepackage{amsmath}
\usepackage{amssymb}
\usepackage{amsfonts}
\usepackage[utf8]{inputenc}
\usepackage{csquotes}
\usepackage{url}
\usepackage{multirow}
\usepackage{subcaption}
\usepackage{array}
\usepackage{float}
\usepackage{hhline}

\usepackage{graphicx}
\usepackage{geometry}
\begin{document}
\title{Scheduling Techniques for Liver Segmentation: ReduceLRonPlateau Vs OneCycleLR\thanks{This publication was made possible by an Award [GSRA6-2-0521-19034] from Qatar National Research Fund (a member of Qatar Foundation). The contents herein are solely the responsibility of the authors. Moreover, the HPC resources and services used in this work were provided by the Research Computing group in Texas A\&M University at Qatar. Research Computing is funded by the Qatar Foundation for Education, Science and Community Development (http://www.qf.org.qa).}}


\titlerunning{ReduceLRonPlateau Vs. OneCycleLR}
%
\author{Ayman Al-Kababji\inst{1}\orcidID{0000-0001-7816-5591} \and
Faycal Bensaali\inst{1}\orcidID{0000-0002-9273-4735} \and
Sarada Prasad Dakua\inst{2}\orcidID{0000-0003-2979-0272}}

\institute{College of Engineering, Qatar University, Doha, Qatar\\ \email{\{aa1405810, f.bensaali\}@qu.edu.qa} \and
Department of Surgery, Hamad Medical Corporation, Doha, Qatar
\email{SDakua@hamad.qa}}

\authorrunning{A. Al-Kababji et al.}


%

%
\maketitle 
\vspace{-1em}
\begin{abstract}
Machine learning and computer vision techniques have influenced many fields including the biomedical one. The aim of this paper is to investigate the important concept of schedulers in manipulating the learning rate (LR), for the liver segmentation task, throughout the training process, focusing on the newly devised \texttt{OneCycleLR} against the \texttt{ReduceLRonPlateau}. A dataset, published in 2018 and produced by the Medical Segmentation Decathlon Challenge organizers, called Task 8 Hepatic Vessel (MSDC-T8) has been used for testing and validation. The reported results that have the same number of maximum epochs (75), and are the average of 5-fold cross-validation, indicate that \texttt{ReduceLRonPlateau} converges faster while maintaining a similar or even better loss score on the validation set when compared to \texttt{OneCycleLR}. The epoch at which the peak LR occurs perhaps should be made early for the \texttt{OneCycleLR} such that the super-convergence feature can be observed. Moreover, the overall results outperform the state-of-the-art results from the researchers who published the liver masks for this dataset. To conclude, both schedulers are suitable for medical segmentation challenges, especially the MSDC-T8 dataset, and can be used confidently in rapidly converging the validation loss with a minimal number of epochs.

\end{abstract}

\section{Introduction}
Hepatic-related diseases are responsible of two million deaths annually around the world \cite{Asrani2019}. Half of these diseases are liver cirrhosis and the other half are hepatitis and hepatocellular carcinoma (HCC) \cite{Asrani2019}. It is also a destination for metastasis originating from adjacent organs. Nonetheless, the liver and its lesions are usually analyzed in tumor staging \cite{Christ2017}.

To aid medical personnel in diseases' diagnosis, it is important to create a delineation for the liver from computerized tomography (CT) and magnetic resonance imaging (MRI) scans, especially for its tumors and vessels. It can help surgeons in pre-procedural planning and/or evaluating the successfulness of a procedure by post-procedural segmentation on a follow-up CT scan. However, at the moment, the norm is to manually or semi-automatically segment the liver as it is more accurate. However, these techniques entail radiologists' subjectivity, intra- and inter-radiologist variance, and time-consumption \cite{Hu2016}.


In surveying related works, U-Net is found to be an important convolutional neural network (ConvNet) architecture that pushed the biomedical segmentation field \cite{Ronneberger2015}. Thus, it was natural for researchers to utilize it extensively. In \cite{Christ2016,Christ2017}, a 2D fully connected network (FCN) following the U-net scheme is utilized, followed by a 3D conditional random field (CRF). In \cite{Zhang2020a}, the U-Net acts as a coarse liver segmenter, but in \cite{Ouhmich2019}, it is used as the main model. On the other hand, other researchers have become inspired from the U-Net structure given its excellence and popularity in different fields. In \cite{Vorontsov2018}, the authors use an ensemble of three 2D FCN models having a U-Net-like architecture for the liver segmentation task, and the final mask is their average. 
In a different direction, \cite{Perslev2019} create a multi-planar U-Net (MPU-Net) to capture the organ of interest from different angles (generalizing to more views than the three conventional ones), and similarly, fusing the output of all planar segmentation to generate the final output. In \cite{Tian2019}, both global and local context are used in a U-Net architecture (GLC-UNet), which also is one of the few studies, if not the first, that attempts to automatically delineate the famous Couinaud segmentation of the liver.


Regarding the scheduling techniques, authors in \cite{Smith2019} discuss a scheduler called \texttt{OneCycleLR} with the idea of regularization via the manipulation of learning rate (LR) to prevent the model from getting trapped at a local minimum well-suited for the training set \cite{Smith2019}. Moreover, super-convergence is achieved as the scheduler allows the model to reach a `better'' minimum for the cost function, for both training and validation set in a shorter time with respect to other scheduling techniques.

The aim of this paper is to compare the \texttt{ReduceLRonPlateau} and \texttt{OneCycleLR} techniques, provided by PyTorch's deep learning library, in training a U-Net ConvNet model for the liver segmentation task. The remainder of this paper is organized as follows. 
Section \ref{sec:methodology} delves into the methodology highlighting the dataset and pre-processing techniques, training environment and the selected parameters, and the test records with the evaluation metrics. Section \ref{sec:results} portrays and discusses the obtained results, and finally, we conclude the paper in Section \ref{sec:conclusion}.

\section{Methodology} \label{sec:methodology}
\begin{figure}[hbtp]
    \centering
    \includegraphics[width=\linewidth, trim={0.95in 2.4in 0.5in 2in}, clip]{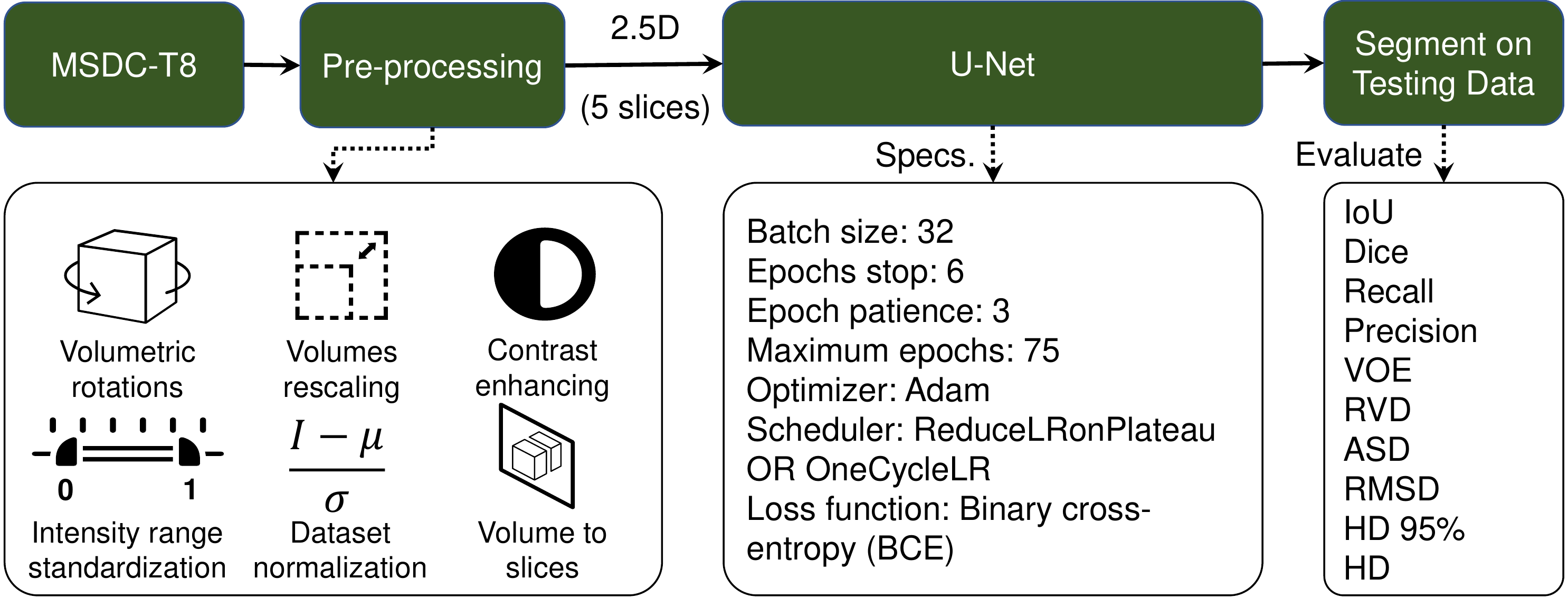}
    \caption{Training and testing scheme}
    \label{fig:methodology}
\end{figure}

In this section, the followed methodology is explained thoroughly. Firstly, the employed dataset is discussed highlighting its characteristics along with the applied pre-processing techniques. Secondly, the utilized model, the environment, and the training parameters are highlighted (no post-processing techniques are applied). Lastly, the test records and the evaluation criteria that are used to evaluate the models' performance are briefly mentioned. Figure~\ref{fig:methodology} highlights the most important steps of training/testing.

\subsection{Dataset and Pre-processing Techniques}
\subsubsection{Dataset Summary}
Out of the 10 produced datasets in The Medical Segmentation Decathlon Challenge, the Task 8 Hepatic Vessels (MSDC-T8) is the one that is used in this paper. The dataset originally contains 443 contrast-enhanced (CE) CT scan records, out of which 303 are for training, and the remaining 140 are for testing. The intra-slice resolution for the whole dataset varies between 0.56 and 0.98 mm, while the inter-slice distance ranges between 0.80 and 8 mm. The slices have the standard number of pixels for a CT scan, which are (512$\times$512), with a varying number of slices between 24 and 251. It is worth noting that the majority, if not all, the records are focused in the abdominal area and contain only the ground-truth labels for both tumors (lesions) and vessels within the livers for the training set. The testing set masks are hidden as part of the challenge. However, the authors in \cite{Tian2019} created the liver annotations within the MSDC-T8 dataset for the whole 443 CT records and shared it publicly, along with Couinaud's segmentation for 193 of those records.

\subsubsection{Pre-processing} The following pre-processing techniques are applied in the order they are mentioned: 1) volumetric rotations so that all volumes face same direction; 2) volumetric rescaling by a half; 3) clipping voxels' intensity values to be in [-100, 400] where the liver is most situated \cite{Yuan2017}; 4) intensity range standardization; 5) contrast enhancing via contrast limited adaptive histogram equalization (CLAHE) technique; 6) data normalization; and finally 7) volumes to slices transformation (3D $\rightarrow$ 2.5D).

\subsection{Training Environment and Parameters}
The used model in this paper is the famous U-Net. It is used to compare the obtained results with the ones reported in \cite{Tian2019}. Moreover, the framework of choice is PyTorch (torch version==1.8.1+cu102) for conducting the comparison, where a 5-fold cross-validation (80\% training/20\% validation) procedure is applied. The maximum number of epochs per fold is 75, but early stopping is utilized to avoid unnecessary training time (epochs\_stop~=~6). The 2.5D (5 slices) input shape is utilized with a batch size~=~32, and binary cross-entropy (BCE) as the loss function. Adam optimizer is used with two different schedulers that are facilitated by the \texttt{torch.optim} library in PyTorch, namely the \texttt{OneCycleLR} \cite{PyTorch-onecyclelr} and the \texttt{ReduceLRonPlateau} schedulers \cite{PyTorch-reducelronplateau}. An NVIDIA Tesla V100 GPU with Intel Xeon Skylake CPU are used for the training.

\subsection{Test Records and Evaluation Metrics}
The 23 testing records were chosen randomly and are mentioned explicitly: [003, 012, 045, 072, 090, 105, 117, 129, 141, 153, 169, 178, 193, 205, 220, 236, 246, 258, 268, 280, 294, 304, 320]. The following metrics are utilized to evaluate the trained models' performance: 1) Dice similarity coefficient (DSC) (per case version) [\%]; 2) intersection-over-union (IoU) [\%]; 3) relative volumetric difference (RVD); 4) average symmetric surface distance (ASD) [mm]; 5) root-mean squared symmetric surface distance (RMSD) [mm]; 6) maximum symmetric surface distance (MSD)/Hausdorff distance (HD) [mm]; 7) 95\% Hausdorff distance (HD95) [mm]; and 8) the number of epochs until early stopping stops the fold run.

\section{Results and Discussion} \label{sec:results}
\begin{figure}[H]
\centering
\begin{subfigure}{.5\textwidth}
  \centering
  \includegraphics[width=\linewidth, trim={0.7in 0.7in 1.3in 0.7in}, clip]{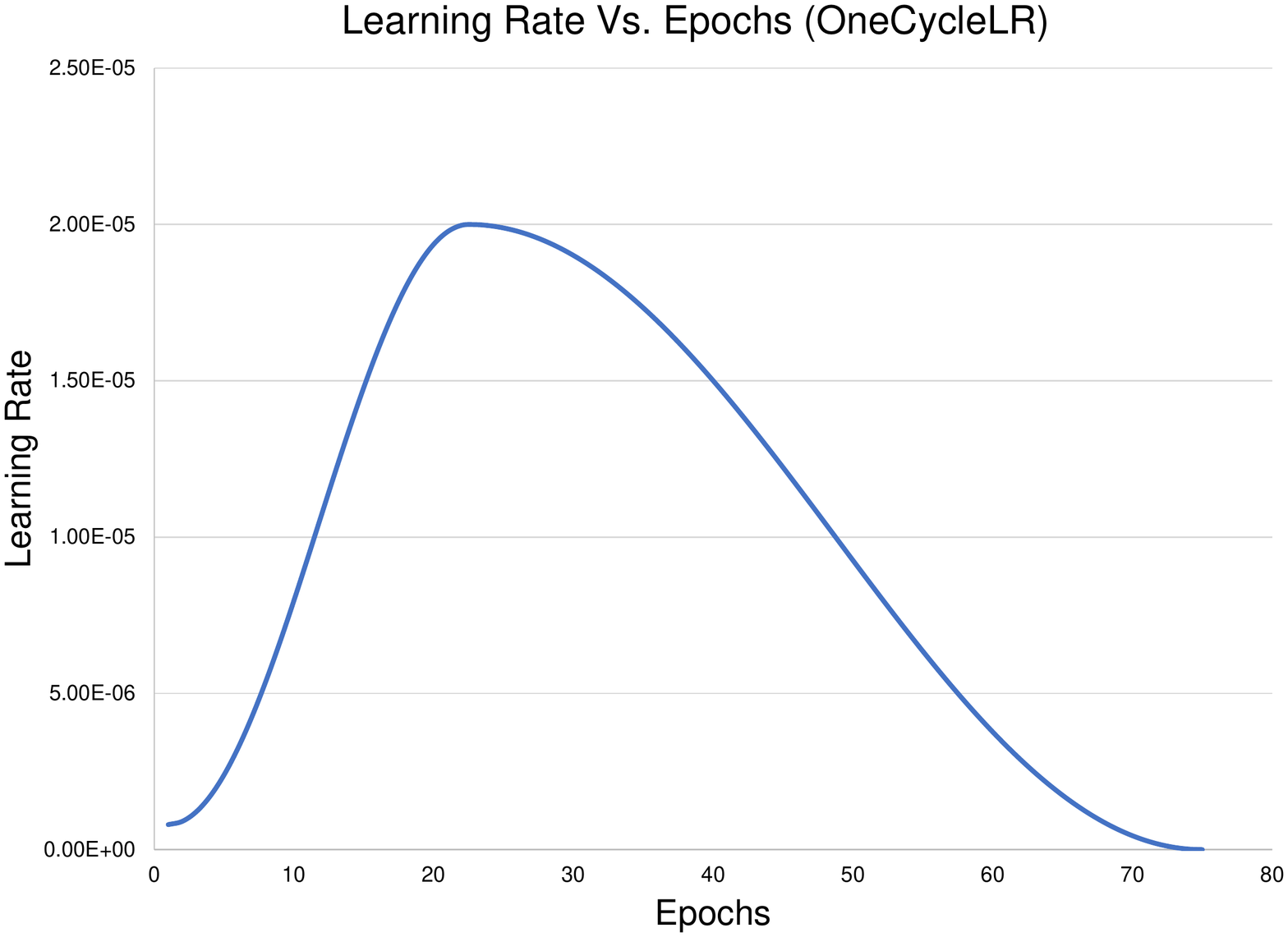}
  \caption{\texttt{OneCycleLR}}
  \label{fig:schedulers_1}
\end{subfigure}%
\begin{subfigure}{.5\textwidth}
  \centering
  \includegraphics[width=\linewidth, trim={0.7in 0.7in 1.3in 0.7in}, clip]{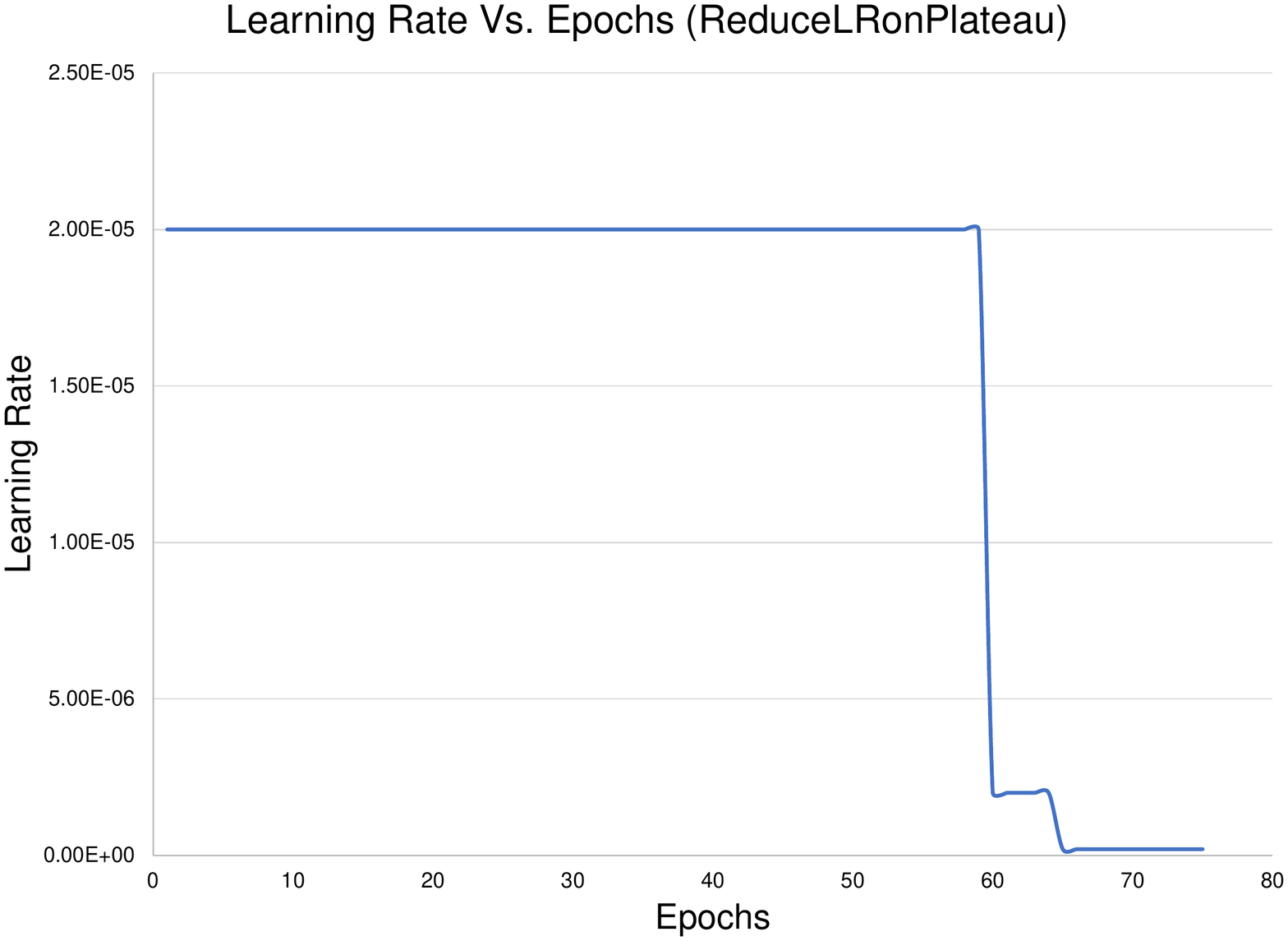}
  \caption{\texttt{ReduceLRonPlateau}}
  \label{fig:schedulers_2}
\end{subfigure}
\caption{Learning rate change using (a) \texttt{OneCycleLR} and (b) \texttt{ReduceLRonPlateau}}
\label{fig:schedulers}
\end{figure}

When implementing the 5-fold cross-validation, $\sim$24K slices end up in training and $\sim$6K slices are in validation. Figure~\ref{fig:schedulers} shows how the LR changes using these two schedulers (assuming the absence of early stopping). The \texttt{ReduceLRonPlateau} scheduler needs at least four parameters to function: 1) initial LR; 2) validation loss to evaluate the model's generalization; 3) epochs\_patience, which is a waiting threshold after which the LR will be changed; and 4) LR\textsubscript{factor} to reduce the LR by a given factor (LR\textsubscript{new} = LR\textsubscript{initial}$\times$LR\textsubscript{factor}). On the other hand, \texttt{OneCycleLR} needs at least two parameters to function: 1) the maximum LR it will peak to; and 2) the number of max\_epochs that is initially planned. 

It can be seen that the \texttt{OneCycleLR} changes the LR irrespective of any metric. On the other hand, \texttt{ReduceLRonPlateau} changes the LR based on saturation or increase in the validation loss. In Figure~\ref{fig:schedulers_1}, the LR starts small and reaches the LR\textsubscript{max} that the user specifies

Due to the variance in how these schedulers operate, different LRs are designed for each scheduler, followed by a discussion drawn over the noticed patterns and results. For both schedulers, seven LRs are selected to showcase the performance of the developed U-Net over the segmentation of the liver. Table~\ref{tab:schedulers_results} shows the mean and standard deviation from the 5-fold cross-validation that is implemented per scheduler and LR.
\vspace{-2.4em}

\begin{table}[H]
\centering
\caption{Results of both schedulers with different LRs are reported through the mean (standard deviation) of 5-fold cross-validation for each metric. Note: The reported LRs for \texttt{OneCycleLR} scheduler are the maximum peaks.}
\begin{tabular}{
|>{\centering\arraybackslash}m{0.5cm}
|>{\centering\arraybackslash}m{1.5cm}
|>{\centering\arraybackslash}m{1.2cm}
|>{\centering\arraybackslash}m{1.2cm}
|>{\centering\arraybackslash}m{1.2cm}
|>{\centering\arraybackslash}m{1.1cm}
|>{\centering\arraybackslash}m{1.3cm}
|>{\centering\arraybackslash}m{1.1cm}
|>{\centering\arraybackslash}m{1.4cm}
|>{\centering\arraybackslash}m{1.3cm}
|}
\hline
& \textbf{LR} & \textbf{Dice} & \textbf{IoU} & \textbf{RVD} & \textbf{ASD} & \textbf{RMSD} & \textbf{HD} & \textbf{95\% HD} & \textbf{Epochs} \\ \hline
\multirow{13}{0.3cm}{\textbf{\rotatebox[origin=c]{90}{OneCycleLR (Max LR)}}} 
            & 2 $\times$10\textsuperscript{-5} & 97.67 (0.12) & 95.48 (0.23) & -0.008 (0.004) & 1.055 (0.509) & 3.81 (1.15) & 44.00 (4.43)  & 5.13 (4.03) & 42.6 (2.6) \\
            & 4 $\times$10\textsuperscript{-5} & 97.72 (0.15) & 95.58 (0.28) & -0.009 (0.004) & 1.002 (0.522) & 3.70 (1.58) & 41.58 (7.41)  & 5.09 (4.12) & 34.0 (3.1) \\
            & 8 $\times$10\textsuperscript{-5} & 97.93 (0.09) & 95.97 (0.17) & -0.009 (0.005) & 0.576 (0.089) & 2.51 (0.30) & 37.78 (1.86)  & 2.82 (0.70) & 29.4 (2.2) \\
            & 16 $\times$10\textsuperscript{-5}  & 98.01 (0.06) & 96.12 (0.11) & -0.008 (0.005) & 0.521 (0.110) & 2.05 (0.45) & 31.51 (6.71)  & 2.68 (0.67) & 29.0 (2.9) \\
            & 24 $\times$10\textsuperscript{-5} & 98.07 (0.11) & 96.24 (0.21) & -0.005 (0.003) & \textbf{0.482 (0.076)} & \textbf{1.88 (0.29)} & 29.25 (5.18) & \textbf{2.43 (0.59)} & 30.0 (3.8) \\
            & 32 $\times$10\textsuperscript{-5} & 98.03 (0.10) & 96.16 (0.19) & \textbf{-0.004 (0.005)} & 0.606 (0.147) & 2.47 (0.71) & 34.31 (3.11) & 2.61 (0.45)  & 25.0 (3.7) \\
            & 40 $\times$10\textsuperscript{-5} & 98.04 (0.17) & 96.17 (0.32) & -0.007 (0.003) & 0.503 (0.131) & 1.97 (0.74) & 33.17 (8.63)  & 2.56 (0.82) & 25.8 (2.0)   \\ \hhline{|=|=|=|=|=|=|=|=|=|=|}
\multirow{13}{0.3cm}{\rotatebox[origin=c]{90}{\textbf{ReduceLRonPlateau}}}
            & 0.4 $\times$10\textsuperscript{-5} & 97.20 (0.13) & 94.60 (0.24) & -0.009 (0.004) & 1.541 (0.276) & 5.59 (0.63) & 59.02 (7.91)  & 10.63 (3.15) & 72.8 (3.9) \\
            & 0.8 $\times$10\textsuperscript{-5} & 97.40 (0.15) & 94.96 (0.27) & -0.010 (0.003) & 1.097 (0.238) & 4.47 (0.86) & 59.47 (7.83)  & 5.56 (3.35)  & 48.6 (4.0) \\
            & 2 $\times$10\textsuperscript{-5} & 97.69 (0.13) & 95.51 (0.24) & -0.011 (0.003) & 0.751 (0.130) & 3.03 (0.61) & 40.43 (1.46)  & 3.40 (0.76)  & 30.6 (1.3) \\
            & 4 $\times$10\textsuperscript{-5} & 97.87 (0.06) & 95.86 (0.11) & -0.011 (0.003) & 0.625 (0.142) & 2.57 (0.84) & 38.26 (7.34)  & 2.73 (0.12)  & 23.0 (2.8) \\
            & 8 $\times$10\textsuperscript{-5} & 98.04 (0.10) & 96.18 (0.19) & -0.009 (0.003) & 0.528 (0.104) & 1.99 (0.52) & \textbf{27.03 (6.55)}  & 2.67 (0.73)  & 20.8 (3.6) \\
            & 16 $\times$10\textsuperscript{-5} & \textbf{98.12 (0.04)} & \textbf{96.33 (0.07)} & -0.008 (0.002) & 0.624 (0.443) & 2.15 (1.40) & 27.16 (4.53)  & 4.10 (4.37)  & 22.0 (1.6) \\
            & 32 $\times$10\textsuperscript{-5} & 98.02 (0.10) & 96.15 (0.19) & -0.006 (0.004) & 0.515 (0.062) & 2.09 (0.52) & 33.87 (10.39) & 2.58 (0.42)  & \textbf{18.8 (4.2)} \\ \hline
\end{tabular} \label{tab:schedulers_results}
\end{table}

\vspace{-1em}
The best results per metrics are bolded in Table~\ref{tab:schedulers_results}. Moreover, The aforementioned LRs for the OneCycleLR are the maximum LRs that the scheduler will reach. Firstly, it is worth noting that the achieved results using the original U-Net in 2.5D mode outperforms the ones described in \cite{Tian2019}. It achieves better Dice result than the 2D (with and without convolutional long short-term memory (LSTM)), 2.5D, and 3D U-Net counterparts in the liver segmentation task. Moreover, it comes really close to the proposed GLC-Unet (98.18 ± 0.85)\% in \cite{Tian2019}, while the best run from \texttt{ReduceLRonPlateau} LR=16$\times$10\textsuperscript{-5} achieves (98.12 ± 0.04)\% in Dice. This shows that it is possible to extract even higher results from their GLC-Unet if they adopt our pre-processing and scheduling techniques. Additionally, the obtained results per fold are close to each other as the reported standard deviation are relatively small.

\begin{figure}[H]
\centering
\begin{subfigure}{.5\textwidth}
  \centering
  \includegraphics[width=\linewidth, trim={0.7in 1.5in 1.25in 1.5in}, clip]{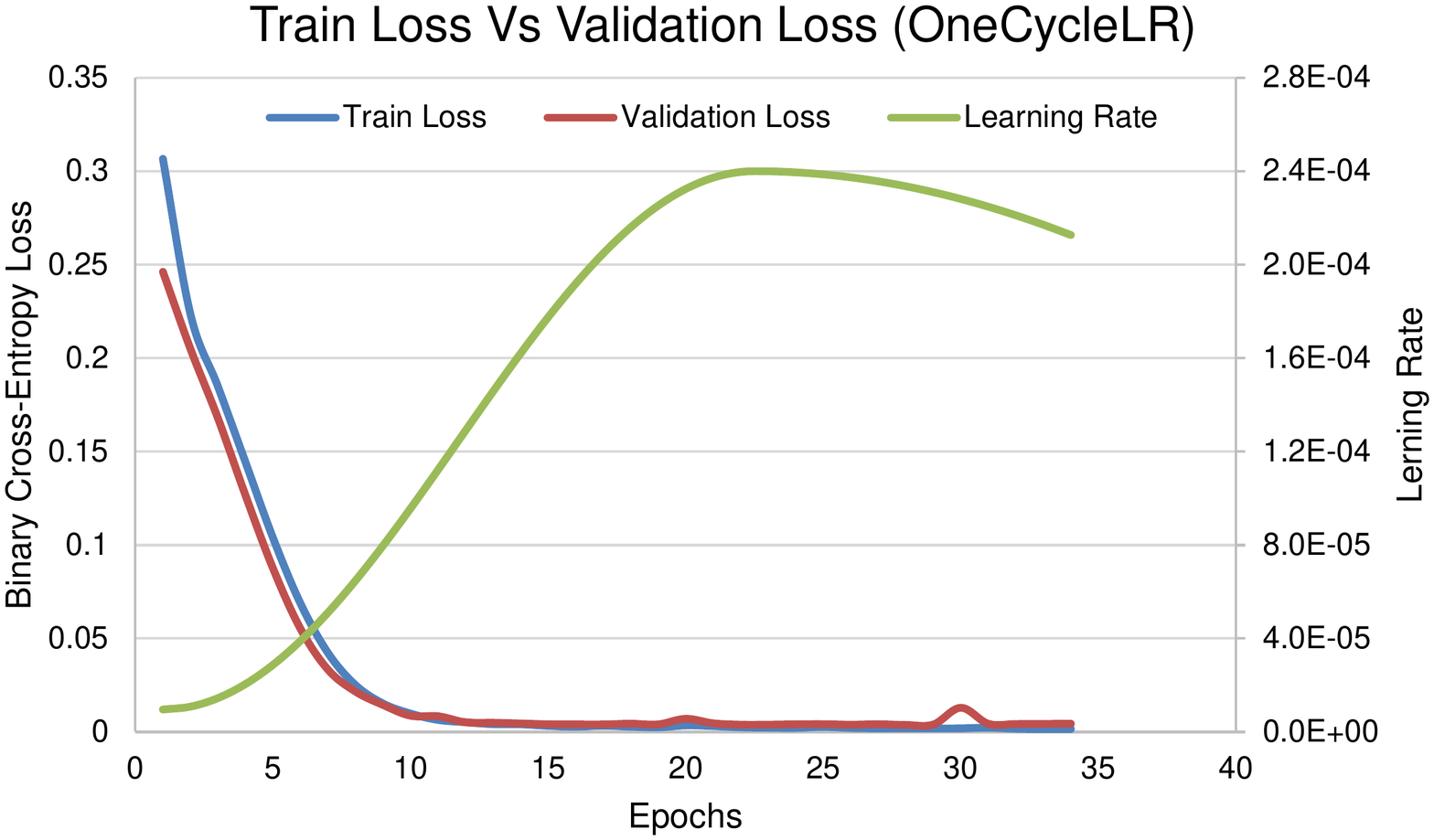}
  \caption{\texttt{OneCycleLR}}
  \label{fig:convergence_schedulers_1}
\end{subfigure}%
\begin{subfigure}{.5\textwidth}
  \centering
  \includegraphics[width=\linewidth, trim={0.7in 1.4in 1.15in 1.5in}, clip]{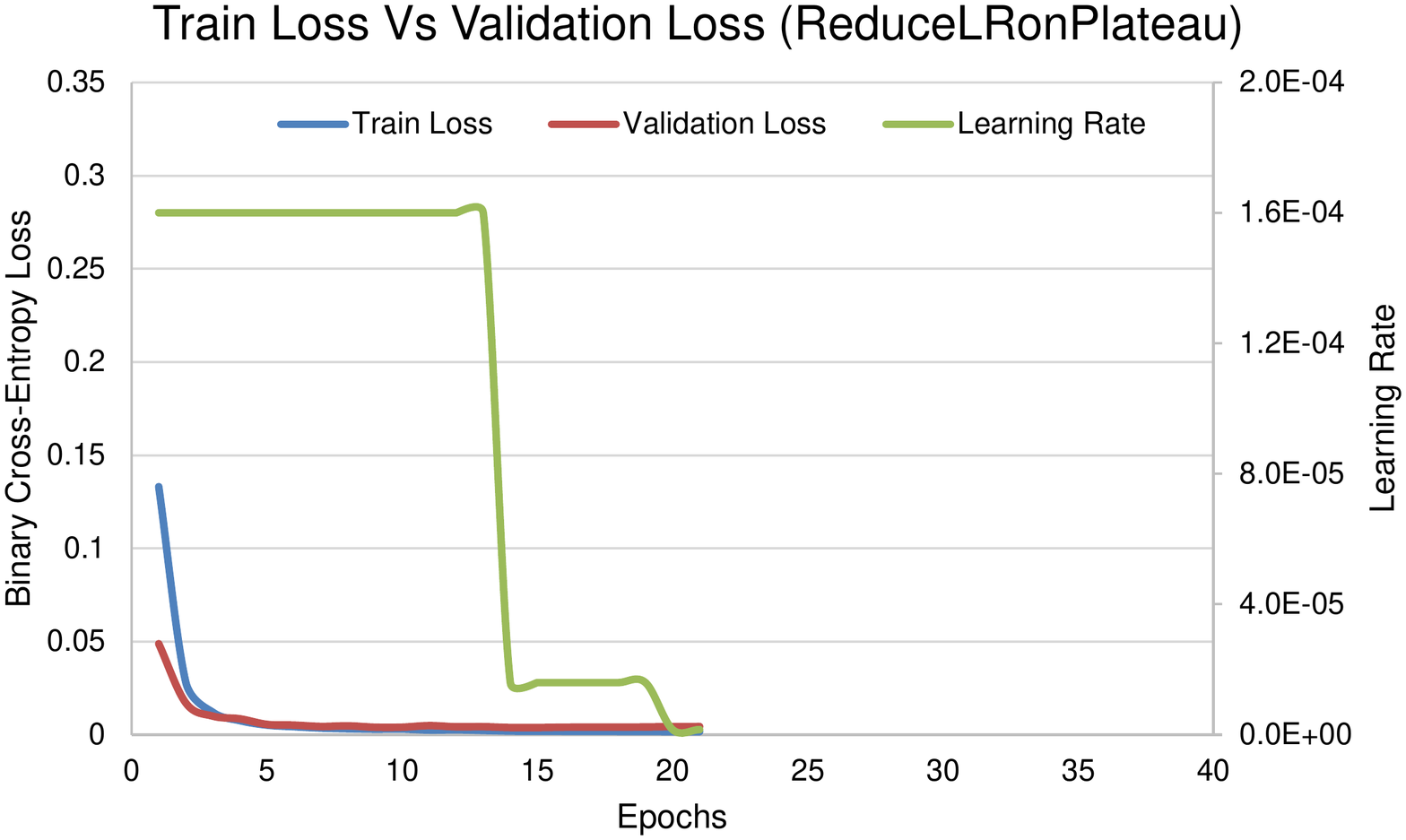}
  \caption{\texttt{ReduceLRonPlateau}}
  \label{fig:convergence_schedulers_2}
\end{subfigure}
\caption{Convergence plots for best fold run (a) \texttt{OneCycleLR} with LR=24$\times$10\textsuperscript{-5} and (b) \texttt{ReduceLRonPlateau} LR=16$\times$10\textsuperscript{-5}}
\label{fig:convergence_schedulers}
\end{figure}

Regarding the chosen LR, it must not be too small such that the network does not get stuck in a local minimum and get prolonged training periods. This is evident for both schedulers, by observing that higher LRs tend to generate ConvNets that generalize better (more fine-tuned) with less training time. In comparing both scheduling techniques in the generated results, the best result from \texttt{ReduceLRonPlateau} (LR=16$\times$10\textsuperscript{-5} with Dice = (98.12 ± 0.04)\%) surpasses the best one from \texttt{OneCycleLR} (LR=24$\times$10\textsuperscript{-5} with Dice = (98.07 ± 0.11)\%), by not a far margin. Moreover, it converged faster than the \texttt{OneCycleLR}. This could be due to the fact that \texttt{OneCycleLR} initially starts with a small LR that gradually increases and reaches the peak around epoch 23 (refer to Figure~\ref{fig:schedulers_1}), while \texttt{ReduceLRonPlateau} starts strong with a high LR with respect to \texttt{OneCycleLR}. Figure~\ref{fig:convergence_schedulers} shows a convergence example from the two best runs from \texttt{OneCycleLR} and \texttt{ReduceLRonPlateau}, respectively, highlighting the aforementioned point on why \texttt{ReduceLRonPlateau} converges faster in this context.

 \vspace{-1em}

\begin{figure}
\centering
\begin{subfigure}{1\textwidth}
  \centering
  \includegraphics[width=\linewidth, trim={0.2in 1.05in 0.2in 0.2in}, clip]{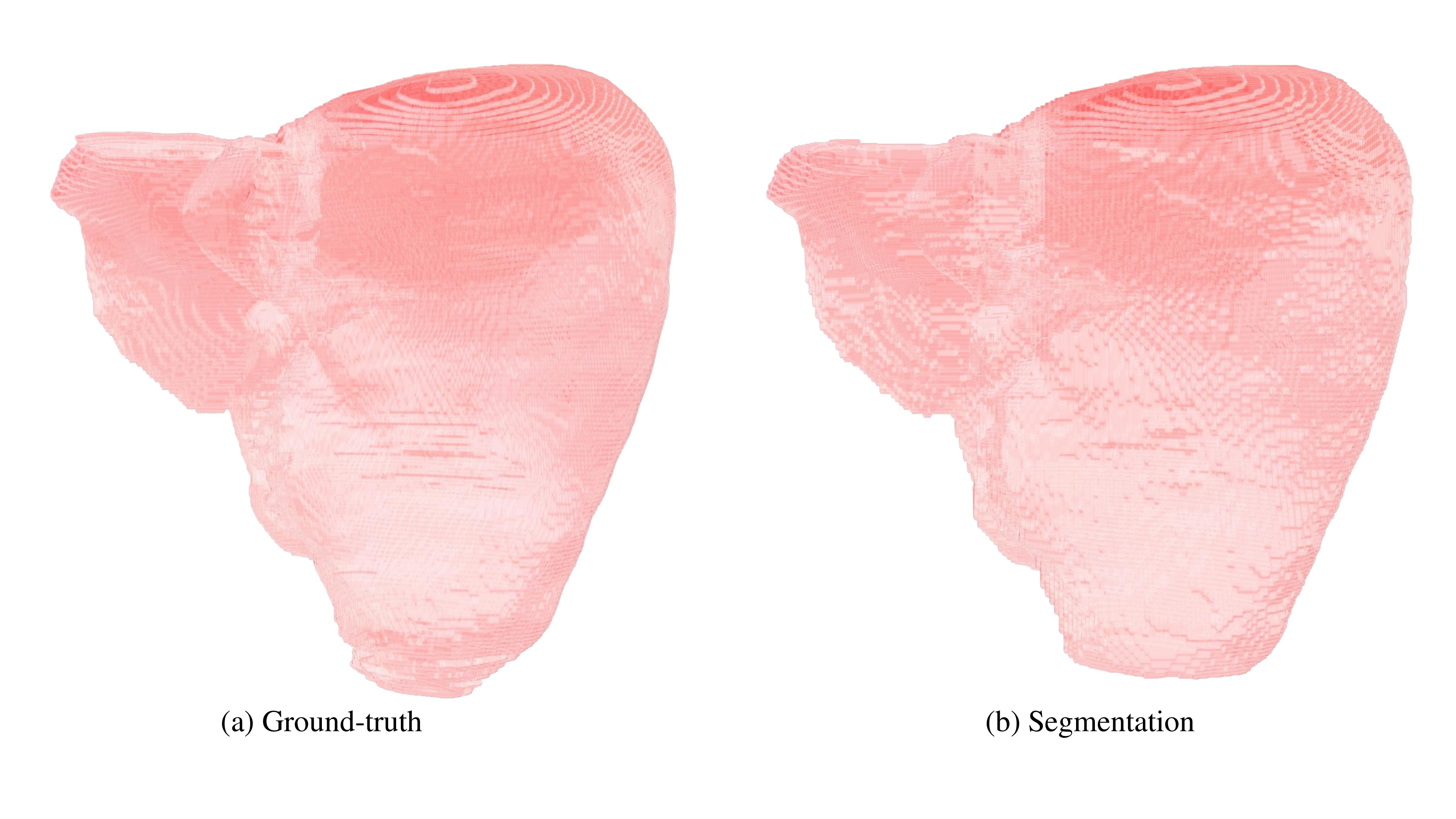}
  \label{fig:3D_construction_0}
\end{subfigure}
\begin{subfigure}{.5\textwidth}
  \centering
  \caption{Ground-Truth}
  \label{fig:3D_construction_1}
\end{subfigure}%
\begin{subfigure}{.5\textwidth}
  \centering
  \caption{Segmentation}
  \label{fig:3D_construction_2}
\end{subfigure}
\caption{Record 294 from MSDC-T8 (a) Ground-Truth and (b) Segmentation (\texttt{ReduceLRonPlateau} LR=16$\times$10\textsuperscript{-5}) fold 5 with Dice = 98.16\%)}
\label{fig:3D_construction}
\end{figure}

After the creation of liver delineations, another script that is developed and built over scikit-image's marching cubes algorithm \cite{scikit-image} is used to build the 3D interpolation of the 2D segmented slices. The script is capable of taking multiple delineations and combining them into a single .obj file along with its .mtl file that will build the liver along with its tumors and vessels in the same .obj file. In this paper, the liver delineations are only present and they are shown in Figure~\ref{fig:3D_construction} for both the ground-truth and segmented masks for record 294 from the MSDC-T8 dataset.

\section{Conclusion and Future Work} \label{sec:conclusion}
To conclude, this paper investigates different scheduling techniques that are aimed towards enhancing the convergence of the trained model. Both schedulers vary the LR to further enhance the model generalization (by making the LR higher at the beginning for the \texttt{OneCycleLR}, and by reducing the LR by a factor for the \texttt{ReduceLRonPlateau}). It is observed that the achieved results outperformed the state-of-the-art results based on U-Net. Moreover, both schedulers performed well on the MSDC-T8 dataset, but \texttt{ReduceLRonPlateau} slightly outperformed  \texttt{OneCycleLR} in this context. For the future work, the aim is to manipulate the number of epochs for the \texttt{OneCycleLR} such that the maximum LR occurs much earlier (prior to epoch 23). Moreover, the tumor and vessels segmentation challenges will be tackled and investigated as well.

\bibliographystyle{splncs04}
\bibliography{main.bib}

\begin{thebibliography}{10}
\providecommand{\url}[1]{\texttt{#1}}
\providecommand{\urlprefix}{URL }
\providecommand{\doi}[1]{https://doi.org/#1}

\bibitem{Asrani2019}
Asrani, S.K., Devarbhavi, H., Eaton, J., Kamath, P.S.: {Burden of liver
  diseases in the world}. J. Hepatol.  \textbf{70}(1),  151--171 (2019).
  \doi{10.1016/j.jhep.2018.09.014}

\bibitem{Christ2016}
Christ, P.F., Elshaer, M.E.A., Ettlinger, F., Tatavarty, S., Bickel, M., Bilic,
  P., Rempfler, M., Armbruster, M., Hofmann, F., D'Anastasi, M., Sommer, W.H.,
  Ahmadi, S.A., Menze, B.H.: {Automatic Liver and Lesion Segmentation in CT
  Using Cascaded Fully Convolutional Neural Networks and 3D Conditional Random
  Fields}. In: Med. Image Comput. Comput. Interv. -- MICCAI 2016. pp. 415--423.
  Springer International Publishing (2016). \doi{10.1007/978-3-319-46723-8\_48}

\bibitem{Christ2017}
Christ, P.F., Ettlinger, F., Gr{\"{u}}n, F., Elshaera, M.E.A., Lipkova, J.,
  Schlecht, S., Ahmaddy, F., Tatavarty, S., Bickel, M., Bilic, P., Rempfler,
  M., Hofmann, F., Anastasi, M.D., Ahmadi, S.A., Kaissis, G., Holch, J.,
  Sommer, W., Braren, R., Heinemann, V., Menze, B.: {Automatic Liver and Tumor
  Segmentation of CT and MRI Volumes using Cascaded Fully Convolutional Neural
  Networks} (2017)

\bibitem{Hu2016}
Hu, P., Wu, F., Peng, J., Liang, P., Kong, D.: {Automatic 3D liver segmentation
  based on deep learning and globally optimized surface evolution}. Phys. Med.
  Biol.  \textbf{61}(24),  8676--8698 (2016).
  \doi{10.1088/1361-6560/61/24/8676}

\bibitem{Ouhmich2019}
Ouhmich, F., Agnus, V., Noblet, V., Heitz, F., Pessaux, P.: {Liver tissue
  segmentation in multiphase CT scans using cascaded convolutional neural
  networks}. Int. J. Comput. Assist. Radiol. Surg.  \textbf{14}(8),  1275--1284
  (2019). \doi{10.1007/s11548-019-01989-z}

\bibitem{Perslev2019}
Perslev, M., Dam, E.B., Pai, A., Igel, C.: {One Network to Segment Them All: A
  General, Lightweight System for Accurate 3D Medical Image Segmentation}. In:
  Med. Image Comput. Comput. Assist. Interv. -- MICCAI 2019. pp. 30--38 (2019).
  \doi{10.1007/978-3-030-32245-8\_4}

\bibitem{PyTorch-onecyclelr}
PyTorch: {OneCycleLR — PyTorch 1.9.0 documentation},
  \url{https://pytorch.org/docs/stable/generated/torch.optim.lr\_scheduler.OneCycleLR.html\#torch.optim.lr\_scheduler.OneCycleLR}

\bibitem{PyTorch-reducelronplateau}
PyTorch: {ReduceLROnPlateau — PyTorch 1.9.0 documentation},
  \url{https://pytorch.org/docs/stable/generated/torch.optim.lr\_scheduler.ReduceLROnPlateau.html\#torch.optim.lr\_scheduler.ReduceLROnPlateau}

\bibitem{Ronneberger2015}
Ronneberger, O., Fischer, P., Brox, T.: {U-net: Convolutional networks for
  biomedical image segmentation}. In: Med. Image Comput. Comput. Assist.
  Interv. -- MICCAI 2015. pp. 234--241. Springer International Publishing
  (2015). \doi{10.1007/978-3-319-24574-4\_28}

\bibitem{Smith2019}
Smith, L.N., Topin, N.: {Super-convergence: very fast training of neural
  networks using large learning rates}. In: Artificial Intelligence and Machine
  Learning for Multi-Domain Operations Applications. vol. 11006. SPIE-Intl Soc
  Optical Eng (may 2019). \doi{10.1117/12.2520589},
  \url{https://www.spiedigitallibrary.org/conference-proceedings-of-spie/11006/1100612/Super-convergence--very-fast-training-of-neural-networks-using/10.1117/12.2520589.short}

\bibitem{Tian2019}
Tian, J., Liu, L., Shi, Z., Xu, F.: {Automatic Couinaud Segmentation from CT
  Volumes on Liver Using GLC-UNet}. In: Med. Image Comput. Comput. Assist.
  Interv. -- MICCAI 2019. pp. 274--282. Springer International Publishing
  (2019). \doi{10.1007/978-3-030-32692-0\_32}

\bibitem{Vorontsov2018}
Vorontsov, E., Tang, A., Pal, C., Kadoury, S.: {Liver lesion segmentation
  informed by joint liver segmentation}. In: 2018 IEEE 15th Int. Symp. Biomed.
  Imaging (ISBI 2018). pp. 1332--1335. IEEE (2018).
  \doi{10.1109/ISBI.2018.8363817}

\bibitem{scikit-image}
van~der Walt, S., {S}ch\"onberger, J.L., {Nunez-Iglesias}, J., {B}oulogne, F.,
  {W}arner, J.D., {Y}ager, N., {G}ouillart, E., {Y}u, T., the scikit-image
  contributors: scikit-image: image processing in {P}ython. PeerJ  \textbf{2},
  ~e453 (6 2014). \doi{10.7717/peerj.453},
  \url{https://doi.org/10.7717/peerj.453}

\bibitem{Yuan2017}
Yuan, Y.: {Hierarchical Convolutional-Deconvolutional Neural Networks for
  Automatic Liver and Tumor Segmentation} (2017)

\bibitem{Zhang2020a}
Zhang, Y., Jiang, B., Wu, J., Ji, D., Liu, Y., Chen, Y., Wu, E.X., Tang, X.:
  {Deep Learning Initialized and Gradient Enhanced Level-Set Based Segmentation
  for Liver Tumor from CT Images}. IEEE Access  \textbf{8},  76056--76068
  (2020). \doi{10.1109/ACCESS.2020.2988647}

\end{thebibliography}

\end{document}